\documentclass[journal]{IEEEtran}
\usepackage{color}
\usepackage{ulem}

\usepackage{tipa}
\def\aref#1{({\ref{#1}})}
\usepackage{cite}

\usepackage[dvipdfmx]{graphicx}

\usepackage{bm} 
\usepackage{epsfig} 
\usepackage{mathptmx} 
\usepackage{times} 
\usepackage{amsmath} 
\usepackage{amssymb}  
\usepackage{algorithm}
\usepackage{algorithmic}

\DeclareMathOperator*{\GEM}{GEM}

\DeclareMathOperator*{\DP}{DP}

\title{\LARGE \bf{Nonparametric Bayesian Double Articulation Analyzer\\ for Direct Language Acquisition from Continuous Speech Signals}}

\author{\thanks{This research was partially supported by a Grant-in-Aid for Young Scientists (B) 2012-2014 (24700233) funded by the Ministry of Education,
Culture, Sports, Science, and Technology, Japan.}%
Tadahiro Taniguchi${}^1$\thanks{${}^1$T. Taniguchi is with College of Information Science and Engineering, Ritsumeikan University, 1-1-1 Noji Higashi, Kusatsu, Shiga 525-8577, Japan {\tt\small  taniguchi@em.ci.ritsumei.ac.jp}}%
, Shogo Nagasaka${}^2$\thanks{${}^1$S. Nagasaka and R. Nakashima are with the Graduate School of Information Science and Engineering, Ritsumeikan University, 1-1-1 Noji Higashi, Kusatsu, Shiga 525-8577, Japan {\tt\small\{ s.nagasaka, nakashima\}@em.ci.ritsumei.ac.jp}}%
, Ryo Nakashima${}^2$%
}

\begin{document}
\maketitle

\begin{abstract}
Human infants can discover words directly from unsegmented speech signals without any explicitly labeled data. 
The main problem of this paper is to develop a computational model that can estimate language and acoustic models, and discover words directly from continuous human speech signals in an unsupervised manner.
 For this purpose, we propose an integrative generative model that combines a language model and an acoustic model into a single generative model called the ``hierarchical Dirichlet process hidden language model'' (HDP-HLM). The HDP-HLM is obtained by extending the hierarchical Dirichlet process hidden semi-Markov model (HDP-HSMM) proposed by Johnson et al. An inference procedure for the HDP-HLM is derived using the blocked Gibbs sampler originally proposed for the HDP-HSMM. This procedure enables the simultaneous and direct inference of language and acoustic models from continuous speech signals. Based on the HDP-HLM and its inference procedure,
we develop a novel machine learning method called nonparametric Bayesian double articulation analyzer (NPB-DAA) that can directly acquire language and acoustic models from observed continuous speech signals.
By assuming HDP-HLM as a generative model of observed time series data, and by inferring latent variables of the model, the method can analyze latent double articulation structure, i.e., hierarchically organized latent words and phonemes, of the data in an unsupervised manner. 
We also carried out two evaluation experiments using synthetic data and actual human continuous speech signals representing Japanese vowel sequences. In the word acquisition and phoneme categorization tasks, the NPB-DAA outperformed a conventional double articulation analyzer (DAA) and baseline automatic speech recognition system whose acoustic model was trained in a supervised manner.  
The main contributions of this paper are as follows: (1) We develop a probabilistic generative model that integrates language and acoustic models, i.e., HDP-HLM. (2) We derive an inference method for this, and propose the NPB-DAA. (3) We show that the NPB-DAA can discover words directly from continuous human speech signals in an unsupervised manner.
\end{abstract}


\IEEEpeerreviewmaketitle

\section{INTRODUCTION}\label{sec1}
\IEEEPARstart{I}{NFANTS} must solve the word segmentation problem in order to acquire language from continuous speech signals to which they are exposed. The word segmentation problem is that of identifying word boundaries in continuous speech. If the speech signals are given to infants as isolated words, the task is easy for them. However, it has been known that a relatively small number of infant-directed utterances consist of an isolated word~\cite{Aslin1996}.
 If infants had knowledge about words and phonemes innately, the problem could be solved relatively easily. 
On the contrary, the fact that each language has different lists of phonemes and words clearly shows that infants have to acquire them through developmental processes. 

From the viewpoint of statistical learning, the learning problem, i.e., direct language acquisition from continuous speech signals, is very difficult because infants do not have access to the truth labels of speech recognition results. In other words, the language acquisition process must be completely unsupervised.The main problem of this paper is to develop a computational model that can estimate language and acoustic models, and discover words directly from continuous human speech signals.

Most modern automatic speech recognition (ASR) systems have a language model that represents knowledge about words and their distributional probabilities as well as an acoustic model that represents knowledge about phonemes and their acoustic features, e.g.,~\cite{Kawahara2000,Dahl2012}. Both are usually trained using large transcribed speech datasets and linguistic corpora through supervised learning. However, infants do not have access to such explicitly labeled datasets.
They have to acquire both language and acoustic models from raw acoustic speech signals in an unsupervised manner. 

The question about what kind of cues human infants utilize to discover words from continuous speech signals arises.
Saffran et al. listed three types of cues for word segmentation: 1)~prosodic, 2)~distributional, and 3)~co-occurrence~\cite{Saffran1996a}.
1)~Prosodic cues rely on acoustic information, such as post-utterance pauses, stressed syllables, and acoustically distinctive final syllables.
2)~Distributional cues represent the statistical relationships between pairs of neighboring speech sounds.
3)~Co-occurrence cues are used by children to learn words by detecting sounds that co-occur with certain entities in the environment.  
Although many researchers had considered the distributional cues to be too complex for infants to use, 
Saffran reported that word segmentation from fluent speech can be accomplished by 8-month-old infants based on solely on distributional cues~\cite{Saffran1996}. It is also reported that the distributional cues seem to be used
by infants  by the age of 7 months, which is earlier than most other cues~\cite{Thiessen2003}.
These results imply that infants have a fundamental mechanism that can estimate word segments using distributional cues.  In addition to this fundamental segmentation mechanism using distributional cues, the prosodic and co-occurrence cues are believed to help the word segmentation task only as 
supplemental cues~\cite{Saffran1996a}. From the viewpoint of phonemic category acquisition, distributional patterns of sounds have been considered to provide infants with clues about the phonemic structure of a language as well~\cite{Kuhl2007}.

Based on these findings, in this paper, we focus on distributional cues. 
We explore the fundamental computational mechanism that can discover words from speech signals using only distributional cues, and develop an unsupervised machine learning method which can discover phonemes and words directly from unsegmented speech signals

In this paper, we propose an unsupervised learning method called the nonparametric Bayesian double articulation analyzer (NPB-DAA) that can automatically estimate double articulation structures, i.e., hierarchically organized latent words and phonemes, embedded in speech signals. We propose this as a computationally valid explanation for the simultaneous acquisition of language and acoustic models. To develop the NPB-DAA, we introduce a probabilistic generative model called the hierarchical Dirichlet process hidden language model (HDP-HLM) as well as its inference algorithm.

The remainder of this paper is organized as follows. 
Section~\ref{sec2} describes the background of the proposed method. 
Section~\ref{sec3} presents the HDP-HLM by extending hierarchical Dirichlet process-hidden semi-Markov model (HDP-HSMM) proposed by Johnson et al.~\cite{johnson2013}. The HDP-HLM is an probabilistic generative model that integrates acoustic and  language models for continuous speech signals. 
Section~\ref{sec4} describes the inference procedure of HDP-HLM, and our proposed NPB-DAA. Sections~\ref{sec5} and \ref{sec6} evaluate the effectiveness of the proposed method using synthetic data and actual sequential vowel speech signals. Section~\ref{sec7} concludes this paper.

\section{Background}\label{sec2}
\subsection{Word segmentation using distributional cues in transcribed data}
With respect to statistical computational models, many kinds of unsupervised machine learning methods for word segmentation have been proposed in the last two decades~\cite{Brent1999,Venkataraman2001,Goldwater2008,Goldwater2009,Mochihashi2009,Johnson2009,Chen2014,Magistry2012,Sakti2011}. 
Brent~\cite{Brent1999} proposed model-based dynamic programming 1 (MBDP-1) for recovering deleted word boundaries in a natural-language text. The MBDP-1 presumes that there is an information source generating the text explicitly and segments the target text so as to maximize the text's probability. Venkataraman~\cite{Venkataraman2001} proposed a statistical model for segmentation and word discovery from phoneme sequences by improving Brent's algorithm.

Recently, Bayesian nonparametrics, including the hierarchical Dirichlet process and hierarchical Pitman-Yor process, have enabled more sophisticated methods for word segmentation. These models have fully Bayesian generative models and make it possible to calculate the appropriately smoothed n-gram probability for a word that has a long context. Theoretically, they can treat an infinite number of possible words. 
Goldwater~\cite{Goldwater2008,Goldwater2009} proposed an HDP-based word segmentation method and showed that taking context into account is important for statistical word segmentation.
Mochihashi et al.~\cite{Mochihashi2009} proposed a nested Pitman-Yor language model (NPYLM), in which a letter n-gram model based on a hierarchical Pitman-Yor language model is embedded in the word n-gram model. They also developed the forward filtering backward sampling procedure to achieve efficient blocked Gibbs sampling and hence infer word boundaries.

However, all of the above mentioned word segmentation methods presume that transcribed phoneme sequences or text data without any recognition errors can be obtained by the learning system. 
In practice, before acquiring a language model containing an inventory of words, a learning system, i.e., an infant,  has to recognize speech signals without any knowledge of words, only with the knowledge of phonemes and/or syllables in an acoustic model. In such a recognition task, the phoneme recognition error rate inevitably becomes high.  
To overcome this problem, several researchers have proposed word discovery methods utilizing co-occurrence cues. 

\subsection{Lexical acquisition using co-occurrence cues}
Roy et al.~\cite{Roy2002a} ambitiously implemented a computational model that enables a robot to autonomously discover words from raw multimodal sensory input. Their results were imperfect compared with recent state-of-art results. However, their results showed it was possible to develop cognitive models that can process raw sensor data and acquire a lexicon without the need for human transcription or labeling.

Iwahashi et al.~\cite{Iwahashi2008} implemented an interactive learning method for a robot to acquire spoken words through human-robot interaction using audio-visual interfaces. Their learning process was carried out on-line, incrementally, actively, and in an unsupervised manner.
Iwahashi et al.~\cite{Iwahashi2003} also proposed a method that enables a robot to learn linguistic knowledge through human-robot communication in an unsupervised manner.
The model combines speech, visual, and behavioral information in a probabilistic framework.
Though its performance was still limited, the model is considered to be a more sophisticated model than that proposed in Roy et al.'s previous study~\cite{Roy2002a} from the viewpoint of statistical machine learning.
On the basis of this work, Iwahashi et al.~\cite{Iwahashi2010} developed an integrated online machine learning system combining speech, visual, and tactile information obtained through interaction. It enabled robots to learn beliefs regarding speech units, words, the concepts of objects, motions, grammar, and pragmatic and communicative capabilities. 
They called the system LCore.

Araki et al.~\cite{Araki2012} built a robot that formed object categories and acquired their names by combining a multimodal latent Dirichlet allocation (MLDA) and the NPYLM. They showed that the iterative learning of MLDA and NPYLM increases word segmentation performance by using distributional cues and co-occurrence cues simultaneously, but they reported that the prediction accuracy decreases as the phoneme recognition error rate increases.
To overcome this problem, Nakamura et al. integrated statistical models for word segmentation and multimodal categorization. They showed that a robot can autonomously form object categories and related words from continuous speech signals and continuous visual, auditory, and haptic information by updating its language and categorization models iteratively~\cite{Nakamura2014}. 

Not only object information, but also place information can be used as co-occurrence cues. 
Taguchi et al.~\cite{Taguchi2011} proposed a method for the unsupervised learning of place-names from information pairs that consist of spoken utterances and the mobile robot's estimated current location without any prior linguistic knowledge other than a phoneme acoustic model. They optimized a word list using a model selection method based on description length criterion.

\subsection{Word segmentation using distributional cues in noisy input}
As described above, it becomes clear that using co-occurrence cues can mitigate the ill effects of phoneme recognition errors in a word discovery task. However, whether or not the word discovery task can be achieved solely from raw speech signals is still an open question.
Neubig et al.~\cite{Neubig2012} extended the unsupervised morphological analyzer proposed by Mochihashi et al. and enabled it to analyze phoneme lattices. Heymann et al.~\cite{Heymann2013} modified Neubig et al.'s algorithm and proposed a suboptimal two-stage algorithm. Heymann et al. reported that their proposed method outperformed the original method in an experiment that used lattice input generated artificially from text input. In addition, they used the discovered language model for phoneme recognition in an iterative manner and reported that recognition performance was improved~\cite{Heymann2014}.
Elsner et al.~\cite{Elsner2013} proposed a computational model that jointly performs word segmentation and learns an explicit model of phonetic variation. However, they did not start with acoustic sound, but with dictated noisy text, i.e., recognized phoneme sequences with errors. Their model does not include acoustic model learning.

\begin{figure*}[t]
  \centering
  \includegraphics[width = 0.8\linewidth]{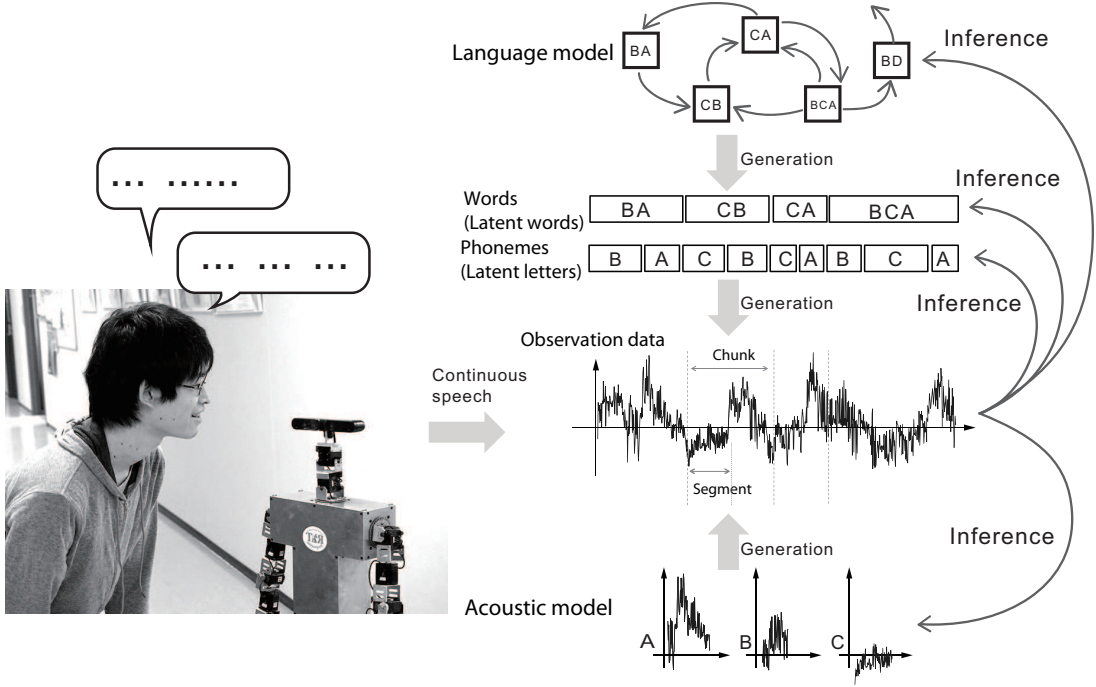}
  \caption{Overview of unsupervised learning of language and acoustic models through human-robot interaction, and the generative process of speech signal assumed in the DAA}
  \label{fig:daa}
\end{figure*}

They showed that the ill effect of phoneme recognition errors can be mitigated to some extent by using distributional information more appropriately.
However, all of these methods, except for Iwahashi et al.,  used an acoustic model previously trained in a supervised manner. Therefore, these models are insufficient as a constructive model for language acquisition from raw speech signals. Hence, the unsupervised learning of an acoustic model is also an important problem. 

\subsection{Unsupervised learning of an acoustic model}

In contrast with the word segmentation task, the acquisition of an acoustic model is basically a categorization task of the feature vectors transformed from continuous speech signals. 
Mixture models, including hidden Markov models (HMMs) and Gaussian mixture models, have been used to model phoneme category acquisition. For example, Lake et al.~\cite{Lake2009} used an online mixture estimation model for vowel category learning. The model was originally proposed by Vallabha et al.~\cite{Vallabha2007}.
However, the phoneme acquisition has proven to be complex categorization task in a feature space. 
The distribution of the feature vectors of each phoneme overlap with each other, and the actual sound of the phoneme depends on its context.
Feldman et al.~\cite{Feldman2013} pointed out that feedback information from segmented words is important for phonetic category acquisition. They demonstrated this effect through simulations using Bayesian models.

Lee et al.~\cite{Lee2012} proposed a hierarchical Bayesian model that can discover a proper set of sub-word units and an acoustic model in an unsupervised manner. However, their model did not estimate the language model.
Lee et al.~\cite{Lee2013a} also proposed a hierarchical Bayesian model simultaneously discovering the phonetic inventory and the Letter-to-Sound mapping rules on the basis of transcribed data only.
The method is not a completely unsupervised learning method from raw speech signals, but does automatically determine relations between sounds and transcribed alphabets and forms an acoustic model in an unsupervised manner.

There have been several studies about the simultaneous unsupervised learning of acoustic and language models.
However, a very small number of statistical learning methods that can simultaneously acquire integrated acoustic and language models  have been proposed. 
Brandl et al.~\cite{Brandl2008} attempted to develop an unsupervised learning method that enables a robot to simultaneously obtain phonemes, syllables, and words from acoustic speech. They did not successfully build such a system, but reported their preliminary results.
Walter et al.~\cite{Walter2013} proposed a word discovery method that uses an HMM-based method for finding acoustic unit descriptors in parallel with a dynamic time warping technique for finding word segments. However, their model is still heuristic from the viewpoint of probabilistic computational models.
As Feldman et al. pointed out, word segmentation and phonetic category acquisition are undoubtedly mutually dependent. Therefore, a theoretically integrated probabilistic generative model for the simultaneous acquisition of language and acoustic models is desirable. 
Very recently, Kamper et al.~\cite{Kamper2015} and Lee et al.~\cite{Lee2015}proposed probabilistic computational models that achieved unsupervised direct word discovery from continuous speech signals. However, they did not provide an explicit, integrated probabilistic generative model for unsupervised simultaneous learning of language and acoustic models.
To develop such an integrated theoretical model, the authors introduced the general concept of double articulation analysis. 

\subsection{Double articulation analysis}

From a general point of view, unsupervised word discovery from raw speech signals is regarded as a double articulation analysis of the time series data representing a speech signal.
The double articulation structure is a well-known two-layer hierarchical structure, i.e., a word sequence is generated from a language model, a word is a sequence of phonemes, and each phoneme outputs observation data during the period it persists.  The word discovery problem becomes a general problem about analyzing the time series data that potentially have a double articulation structure by estimating the latent acoustic model as well as the latent language model.

Taniguchi et al.~\cite{Taniguchi2011} proposed a double articulation analyzer (DAA) by combining the sticky HDP-HMM and the NPYLM. The sticky HDP-HMM proposed by Fox et al. is an nonparametric Bayesian extension of HMM~\cite{Fox2009}.
They applied the DAA to human motion data to extract unit motion from unsegmented human motion data.
However, they simply used the two nonparametric Bayesian methods sequentially. They did not integrate the two models into a single generative model. Therefore, if there are many recognition or categorization errors in the result of the first latent letter recognition process, i.e., segmentation process by the sticky HDP-HMM, the performance of the subsequent process, i.e., unsupervised chunking by the NPYLM, deteriorates. 
In the terminology of a DAA, a latent letter and a latent word basically correspond to a phoneme and a word in speech signals, respectively.
In this paper, we call this method ``conventional DAA'' in order to differentiate it from the DAA newly proposed in this paper, i.e., NPB-DAA. Conventional DAA has been successfully applied to human motion data and driving behavior data, which were also considered to potentially have a double articulation structure. Conventional DAA has been used for various purposes, e.g., segmentation~\cite{takenaka12iros}, prediction~\cite{taniguchi12iv,IEEE-ITS}, data mining~\cite{genki12sii}, topic modeling~\cite{bando13iv,bando13iros}, and video summarization~\cite{takenaka12mm}.
Conventional DAA owes its successful result with respect to driving behavior data to the fact that driving behavior data were continuous and smooth compared with raw speech signals. For a driving letter, which corresponds to a phoneme in continuous speech signals, the recognition error rate was still low. 
However, it is expected that a straightforward application of the conventional DAA to raw speech signals will inevitably turn out badly.

Therefore, based on the background mentioned above, in this paper, we propose an integrated probabilistic generative model, HDP-HLM, representing a latent double articulation structure that contains both a language model and an acoustic model. By assuming HDP-HLM as a generative model of observed time series data, and by inferring latent variables of the model, we can analyze latent double articulation structure of the data in an unsupervised manner. A novel double articulation analyzer is developed on the basis of the HDP-HLM and its inference algorithm.
This HDP-HLM-based double articulation analysis method is called NPB-DAA.

\section{Generative model}\label{sec3}
In this section, we propose a novel generative model, the HDP-HLM, for time series data  that potentially has a double articulation structure, by extending HDP-HSMM~\cite{johnson2013}. 
As indicated in its name, HDP-HLM latently contains a language model. In contrast with the conventional case where a latent state transits to the next state on the basis of a Markov process in the HDP-HMM, a latent word in the HDP-HLM transits to the next latent word on the basis of a language model. 
An illustrative overview of the proposed method and the target task are shown in Fig.~\ref{fig:daa}.
We can naturally derive an inference procedure for the HDP-HLM based on the blocked Gibbs sampler. First, we briefly describe the HDP-HSMM. We then describe the HDP-HLM.

\subsection{HDP-HSMM}
\begin{figure}[t]
  \centering
  \includegraphics[width = 1.0\linewidth]{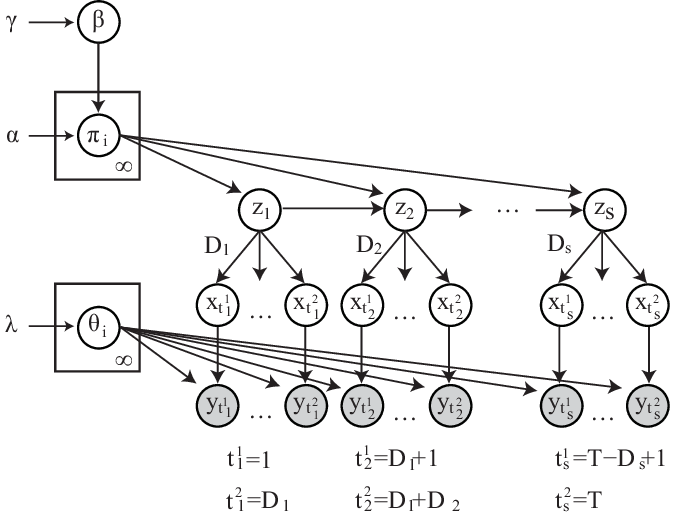}
  \caption{Model of the HDP-HSMM~\cite{johnson2013}}
  \label{fig:graphical_model_of_hdp-hmm}
\end{figure}

HDP-HSMM is a nonparametric Bayesian extension of the conventional hidden semi-Markov model (HSMM)~\cite{johnson2013,HSMM}. Unlike HDP-HMM, which is an nonparametric Bayesian extension of conventional hidden Markov model (HMM)~\cite{Fox2009,teh2006}, the HDP-HSMM explicitly models the duration time of a hidden state. 
A graphical model of the HDP-HSMM is shown in Fig.~\ref{fig:graphical_model_of_hdp-hmm}.
The generative process of the HDP-HSMM is described as follows. 
\begin{align}
  \beta &\sim \GEM(\gamma)\\
  \pi_i &\sim \DP(\alpha, \beta) && i = 1, 2, \dots, \infty\\
  (\theta_i, \omega_i) &\sim H \times G && i = 1, 2, \dots, \infty\\
  z_s &\sim \pi_{z_{s-1}} && s = 1, 2, \dots, S\\
  D_s &\sim g(\omega_{z_{s}})\\
  x_t &= z_s && t = t_s^1,t_s^1 +1,  \dots, t_s^2 \\
  y_t &= h(\theta_{x_t})\\
  t_s^1 &= \sum_{s' < s} D_{s'}\\
  t_s^2 &= t_s^1 + D_s - 1
\end{align}
where $\GEM$ and $\DP$ represent the stick breaking process and Dirichlet process, respectively~\cite{Sethuraman1994,teh2006}.
The parameters $\gamma$ and $\alpha$ are hyperparameters of the $\DP$, $\beta$ is a global transition probability that becomes the base measure of the transition probability distributions, and $\pi_i$ is a transition probability distribution related to the $i$-th super state. Variable {$z_s$} is the $s$-th super state in the sequence of super states, {$D_s$} is the frame duration of {$z_s$},
and the variables {$x_t$} and {$y_t$} are a hidden state and an observation at time frame {$t$}, respectively.
Parameters of an emission distribution and a duration distribution for the {$i$}-th super state are described as 
{$\theta_{i}$} and {$\omega_i$}. Additionally, $H$ and $G$ are base measures for emission distribution and duration distribution. The function $h$ and $g$ represent emission and duration distributions, respectively. The time frames {$t^1_s$} and {$t^2_s$} are frames corresponding to a start point and a end point of a segment corresponding to $z_s$.

In contrast with the case where HMM assumes that a hidden state $x_t$ transits to the next hidden state $x_{t+1}$ according to a Markov process, the hidden semi-Markov Model (HSMM) assumes that a hidden super state $z_s$ transits to next hidden super state $z_{s+1}$ after a probabilistically determined duration time $D_s$, which is sampled from a duration distribution $g(\omega_{z_s})$ 
The super state $z_s$ is sampled from a categorical distribution $\pi_{z_{s-1}}$ related to the previous super state $z_{s -1}$.
When the super state $z_s$ and duration time $D_s$ are sampled, a sequence of hidden states $\{ x_t \mid  1+\sum_{s'=1}^{s-1}D_{s'} \le t \le \sum_{s'=1}^{s}D_{s'}  \}$ are determined to be $z_s$. 

An observation datum $y_t$ at time $t$ is assumed to be drawn from an emission distribution $h$ whose parameter is $\theta_{x_t}$. 
Observation data $y_t$ are generated by $h(\theta_{x_t})$ for $D_s$ steps.

An efficient sampling inference procedure based on the backward filtering forward sampling technique was proposed for 
constructing a blocked Gibbs sampler~\cite{johnson2013}. A similar algorithm was proposed for HDP-HMM by Fox et al.~\cite{Fox2009}. The algorithm is derived from a weak-limit approximation of the number of hidden super states. The computational cost of the message passing algorithm can be reduced to $O(Td_{\max}N^2)$, where $T$ is the length of the observed data, $N$ is the state cardinality, and $d_{\max}$ is the maximal duration of a super state for truncation. The order is almost the same as that of the backward filtering forward sampling algorithm for the HDP-HMM, except for the constant factor $d_{\max}$.

\subsection{HDP-HLM}
\begin{figure*}[t]
  \centering
  \includegraphics[width = 0.7\linewidth]{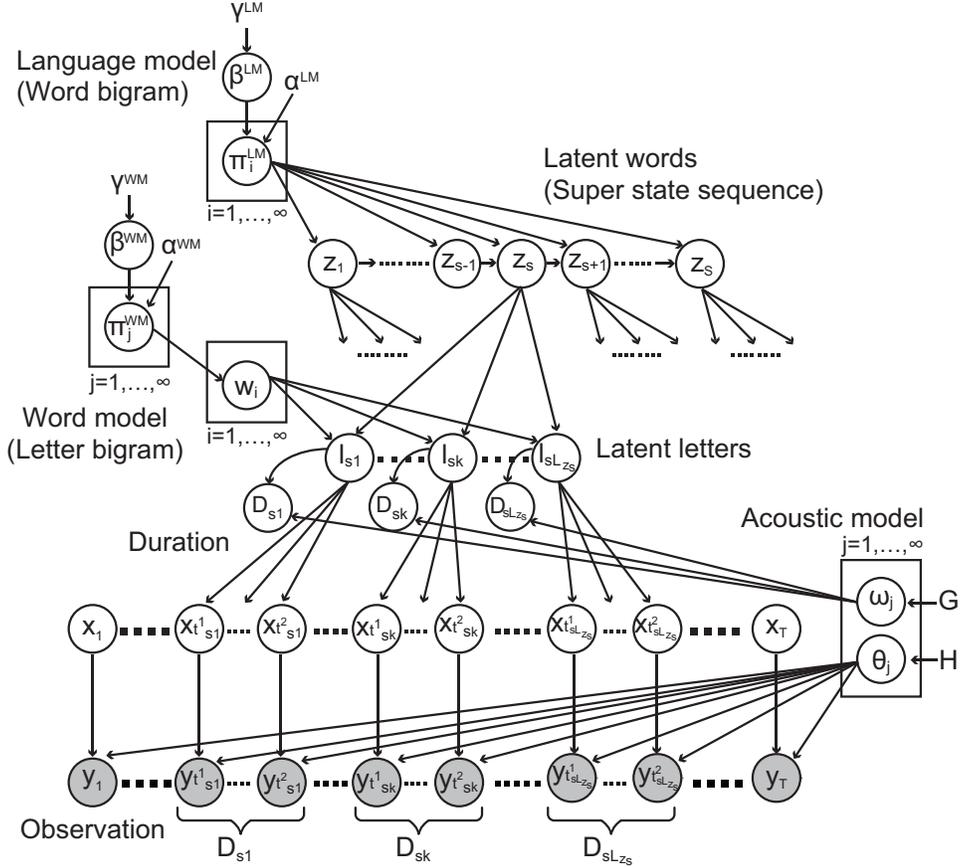}
  \caption{Model of the proposed HDP-HLM}
  \label{fig:graphical_model}
\end{figure*}

The generative model for time series data that potentially have a double articulation structure can be obtained by extending the HDP-HSMM. 
A graphical model of the proposed HDP-HLM is shown in Fig.~\ref{fig:graphical_model}.
In the generative model of HDP-HLM, the super state $z_s$ corresponds to a word in spoken language, which is the fundamental idea of the extension. 
 The $i$-th super state $z_s = i$ has a phoneme sequence $w_{i} = (w_{i1}, \ldots, w_{ik}, \ldots, w_{iL_i})$, where $L_i$ is the length of the $i$-th word $w_i$. 
 The generative process of the HDP-HLM is described as follows. 
\begin{align} 
  \beta^{LM} &\sim \GEM(\gamma^{LM})\\
  \pi^{LM}_i &\sim \DP(\alpha^{LM}, \beta^{LM}) && i = 1, 2, \dots, \infty
  \end{align}
  \begin{align}
  \beta^{WM} &\sim \GEM(\gamma^{WM})\\
  \pi_j^{WM} &\sim \DP(\alpha^{WM}, \beta^{WM}) && j = 1, 2, \dots, \infty
\end{align}
\begin{align}
  w_{ik} &\sim \pi_{w_{ik-1}}^{WM} && i = 1, 2, \dots, \infty,\nonumber\\&&& k = 1, 2, \dots, L_{i}\\
  (\theta_j, \omega_j) &\sim H \times G && j = 1, 2, \dots, \infty
\end{align}
\begin{align}
  z_s &\sim \pi^{LM}_{z_{s-1}} && s = 1, 2, \dots, S\\
  l_{sk} &= w_{z_sk} && s = 1, 2, \dots, S\\
  &&& k = 1, 2, \dots, L_{z_s}\\
  D_{sk} &\sim g(\omega_{l_{sk}}) && s = 1, 2, \dots, S\\
  &&& k = 1, 2, \dots, L_{z_s}
\end{align}
\begin{align}
 x_t &= l_{sk} && t = t_{sk}^1 , \ldots , t_{sk}^2\\
 &&& t_{sk}^1 = \sum_{s' < s} D_{s'} + \sum_{k' < k} D_{sk'} +1 \\
 &&& t_{sk}^2 = t_{sk}^1 + D_{sk} - 1\\
  y_t &= h(\theta_{x_t}) && t = 1, 2,  \ldots , T
\end{align} 
where $\beta^{WM}$ is the base measure and $\alpha^{WM}$ and $\gamma^{WM}$ are hyperparameters of a word model, which generates words, i.e., latent letter sequences. 
Furthermore, $\DP (\alpha^{WM}, \beta^{WM})$ outputs $\pi^{WM}_j$, representing the transition probability from latent letter $j$ to the next latent letter. By contrast, $\beta^{LM}$ is the base measure, $\alpha^{LM}$ and $\gamma^{LM}$ are hyperparameters of the language model,  and $\DP(\alpha^{LM}, \beta^{LM})$ outputs $\pi^{LM}_i$, representing the transition probability from latent word $i$ to the next latent word. The superscripts $LM$ and $WM$ indicate language model (LM) or word model (WM), respectively. 
The latent letters contained in the $i$-th latent word $w_i$ are sequentially sampled from $\pi^{WM}_{ w_{ik - 1}}$. The $k$-th latent letter of the $i$-th latent word is represented by $w_{ik}$. 
The emission distribution {$h$} and the duration distribution {$g$} have parameters {$\theta_j$} and {$\omega_j$} for the $j$-th latent letter, respectively. The base measures {$H$} and {$G$} generate {$\theta_j$} and {$w_j$}, respectively. Variable {$z_s$} is the $s$-th latent word in the sequence of latent words, and corresponds to the super state in HDP-HSMM, {$D_s$} is the frame duration of {$z_s$}, $l_{sk} = w_{z_sk}$ is the $k$-th latent letter of the $s$-th latent word, and {$D_{sk}$} is the frame duration of {$l_{sk}$}. The variable {$x_t$} and {$y_t$} are a hidden state and an observation at time frame {$t$}, rspectively. The time frames {$t^1_{sk}$} and {$t^2_{sk}$} are frames corresponding to a start point and a end point of a segment corresponding to {$l_{sk}$}, respectively.

In contrast with HMMs, the duration distribution is explicitly determined for each latent letter $l_{sk}$ in the HDP-HLM. The HDP-HLM inherits this property from the HDP-HSMM~\cite{johnson2013}. The duration time $D_{sk}$ of latent letter $l_{sk}$, which is the $k$-th latent letter of the $s$-th latent word $z_s$ in a sampled word sequence, is drawn from the duration distribution $g(\omega_{l_{sk}})$, where $\omega_{l_{sk}}$ is the duration parameter for latent letter $l_{sk}$.
The duration of a latent word $w_{z_s}$ becomes $D_s = \sum_{k = 1}^{L_{z_s}} D_{sk}$. If we assume that {$g$} is a Poisson distribution, the duration distribution of a latent word {$z_s$} also follows a Poisson distribution. In this case, the Poisson parameter of the duration distribution becomes {$\sum_{k = 1}^{L_{z_s}} \omega_{l_{sk}}$}. This relation owes to the reproductive property of Poisson distributions.

In the HDP-HLM,  latent word $z_s$ determines a latent letter sequence $l_{sk} = w_{{z_s}k} \ (k = 1, 2, \ldots , L_{z_s})$. Based on the determined sequence $w_{z_s}$,  duration $D_{sk}$ of $l_{sk}$ is drawn, and observations $y_t$ are drawn from an emission distribution $h(\theta_{x_t})$ corresponding to $x_t = l_{s(t)k(t)}$. The maps $s(t)$ and $k(t)$ represent the indices of words and letters, respectively, in a latent word sequence at time $t$. Using this generative model, a continuous time series data with a latent double articulation structure can be generated.
In this paper, we assume that observed time series data $y_t$ represents a feature vector of the speech signal at time $t$ and is generated in this way. Generally, the HDP-HLM can be applied to any kind of time series data that has a double articulation structure. 

From the viewpoint of language acquisition, we review the generative model. In the conventional DAA~\cite{Taniguchi2011}, a DAA is composed of two separated machine learning methods, i.e., sticky HDP-HMM for encoding observation data to letter sequences and NPYLM for chunking letter sequences into word sequences.  
On the one hand, the transition probabilities $\pi_i^{LM}$ and $\pi ^{WM}_i $ correspond to the word bigram and letter bigram models in the NPYLM, respectively. Therefore, $(\pi^{LM}, \pi ^{WM})$ contains information regarding a language model. On the other hand, $\{ \omega_j, \theta_j \}_{j=1,2, \ldots , \infty}$ contains information regarding an acoustic model, which corresponds to a sticky HDP-HMM in conventional DAA. 

The HDP-HLM assumes that the language model consists of a word bigram model. Mochihashi et al. compared the bigram and trigram language models and showed that the trigram assumption hardly improved the word segmentation performance although computational cost and complexity increased~\cite{Mochihashi2009}. Therefore, the bigram assumption must be appropriate for a word segmentation and word discovery task.

If we derive an efficient inference procedure for this two-layer hierarchical generative model, the inference procedure can infer the acoustic model and language model simultaneously.

\section{Inference algorithm}\label{sec4}
In this section, we derive an approximated blocked Gibbs sampler for the HDP-HLM. The sampler can simultaneously infer latent letters, latent words, a language model, and an acoustic model. Concurrently, the inference procedure can estimate the overall double articulation structure from continuous time series data. Therefore, we propose the unsupervised machine learning method NPB-DAA.
The overall inference procedure is shown in Algorithm 1.
%

\subsection{Inference of latent words: $z_s$}
In the HDP-HSMM, a backward filtering forward sampling procedure is adopted instead of the direct assignment procedure. When each latent state strongly depends on other neighboring latent states, the direct assignment procedure, which is  a naive implementation of the Gibbs sampler, results in a poor mixing rate~\cite{johnson2013}. 
Johnson et al. showed that a blocked Gibbs sampler using a backward filtering forward sampling procedure that can simultaneously sample all hidden states of an observed sequence outperforms a direct-assignment Gibbs sampler. 
By extending the backward filtering forward-sampling procedure and making it applicable to HDP-HLM, we can obtain an inference procedure for HDP-HLM.

The calculation of the backward messages for super states $i$ in HDP-HSMM is as follows.
\begin{align}
  B_t(i) & = P(y_{t + 1: T} \mid z_{s(t)} = i, F_t = 1)\label{eq:1}\\
  &= \sum_j B_t^*(j)P(z_{s(t + 1)} = j \mid z_{s(t)} = i)\label{eq:3}\\
  B_t^{*}(i) &= P(y_{t + 1 : T} \mid z_{s(t + 1)} = i, F_t = 1)\label{eq:2}\\
  &= \sum_{d = 1}^{T-t}B_{t + d}(i)P(D_{t+1}=d \mid z_{s(t + 1)} = i) \nonumber \\
  & ~~~~\times P(y_{t + 1 : t + d} \mid z_{s(t + 1)} = i, D_{t + 1} = d)\label{eq:4}\\
  B_T(i) &= 1\label{eq:9}
\end{align}
where $F_t$ is a variable indicating that $t$ is the boundary of the super state. If $F_t = 1$, $z_{s(t)} \ne z_{s(t + 1)}$. The variable $B_t(i)$ in \aref{eq:1} represents the probability that the latent super state $z_{s(t)} = i$ and that it transitions into a different super state at the next time step. Probability $B_t(i)$ is obtained by marginalizing over all super states $j$ at time step $t + 1$. Variable $B_t^*(j)$ in \aref{eq:2} represents the probability that the latent super state becomes $j$ from time step $t + 1$. This probability can be obtained by marginalizing over the duration variable in \aref{eq:4}. Probability $P(y_{t + 1:t + d} \mid z_{s(t + 1)}=i, D_{t + 1} = d)$ in  \aref{eq:4} shows the emission probability of observed data $y_{t + 1:t + d} $ given the condition that the duration $D_{t + 1}$ of $z_{s(t + 1)}$ is $d$. 
In the HDP-HSMM, all time steps with the same super state $z$ share the same emission distribution. Therefore, the likelihood of a super state $z_{s(t+1)}$, i.e., $P(y_{t + 1 : t + d} \mid z_{t + 1}, D_{t + 1} = d)$, can be calculated easily. 

Surprisingly, in HDP-HLM, the exact same procedure of calculating backward messages as that of HDP-HSMM can be used. 
We obtain a message passing algorithm for HDP-HLM by replacing a super state $z_s$ in HDP-HSMM with latent word $z_s$ in HDP-HLM. Only the likelihood of the latent word $w_s$, i.e., $P(y_{t + 1:t + d} \mid z_{s(t + 1)}=i, D_{t + 1} = d)$, is different between the two message passing algorithms.
 The likelihood of the occurrence of latent word $z_{s(t + 1)}=i$ then becomes
\begin{align}
  P(y_{t + 1 : t + d}&\mid z_{s(t+1)} = i, D_{t+1}=d) \nonumber \\
 & =  \sum_{r \in R^{(L_i, d)}}  \prod_{k = 1}^{L_i}
             P(r_k \mid \omega_{w_{ik}})\times  \nonumber \\
             &~~~~~~~~~~\prod_{m = 1}^{r_k} {P(y_{t + m + \sum_{k'=1}^{k-1}r_{k'}} \mid \theta_{w_{ik}})}. \label{eq:5}\\
   R^{(L_i,d)}&= \left \{r \mid |r| = L_i, \sum_{k =  1}^{|r|} r_k = d \right \}
\end{align}
where $|x|$ indicates the number of elements in vector $x$, and $r = (r_i, r_2, \ldots , r_{L_i})$ is an $L_i$-partition of duration $d$.
By substituting \aref{eq:5} into \aref{eq:4}, we can obtain a formula to calculate the backward message of HDP-HLM. 

The calculation of \aref{eq:5} looks complicated at first glance. However, the 
value of \aref{eq:5} can be efficiently calculated using dynamic programming. If we define forward message $\alpha_{t}(k)$ as the probability that the $k$-th latent letter in the relevant latent word $w_i$ transits to the next latent letter at time $t$ after emitting observations, forward message $\alpha_t(k)$ can be recursively calculated as follows:
\begin{align}
  \alpha_{t}(k) &= \sum_{d' = 1}^{t - k + 1} \alpha_{t - d'}(k - 1) P(d' \mid \omega_{w_{ik}})
  \prod_{t' = 0} ^{d' - 1} P(y_{t - t'} \mid \theta_{w_{ik}}) \\
  \alpha_0(0) &= 1.
\end{align}
As a result, $P(y_{t + 1 : t + d} \mid z_{s(t+1)} = i, D_{t+1}=d) = \alpha_d(L_i)$.
By applying the calculation formula shown above, backward messages $B_t(i)$ and $B_t^*(i)$ can be calculated. Using the calculation procedure for backward messages, the forward sampling procedure proposed in the HDP-HSMM can be employed. The backward filtering forward sampling procedure enables the blocked Gibbs sampler to directly sample latent words from observation data without explicitly sampling latent letters in HDP-HLM.

In the forward sampling procedure, super state $z_s$ and its duration $D_s$ are sampled iteratively using backward messages as follows. 
\begin{align}
&P(z_s = i | y_{1:T}, z_{s-1} = j, F_{D^\text{sum}_s}=1) = \nonumber\\
&~~~~P(z_s = i| z_{s-1} = j) B_{D^\text{sum}_s}(i)P(y_{D^\text{sum}_s} | z_s = i)\\
&P(D_s = d | y_{1:T}, z_{s} = i,F_{D^\text{sum}_s}=1  ) = P(D_s = d)\times\nonumber\\
&~~~~\frac{P(y_{D^\text{sum}_s+1 : D^\text{sum}_s + d}| D_s = d, z_s=i ,F_{D^\text{sum}_s}=1 ) B_{D^\text{sum}_s + d}(i)}{B^*_{D^\text{sum}_s}(i)}
\end{align}
where $D^\text{sum}_s = \sum_{s'<s}D_{s'}$. 
For further details, please refer to the original paper, in which the HDP-HSMM was introduced~\cite{johnson2013}.

\subsection{Sampling a letter sequence for a latent word: $w_i$}
The sampled $z_s$ is only an index of a latent word. Concrete letter sequences $w_i$ for each latent word $i$ should be sampled according to the correspondence of each sub-sequence of time series data $\bm{y}^k = (y^k_1, y^k_2, \ldots , y^k_{T^k})$ to each latent word. 
When a latent word $z_s$ is given, the generative model of the observation in the range of a latent word $z_s$ can be regarded as an HDP-HSMM whose super states correspond to latent letters. Therefore, in the proposed model, each sub-sequence of observation data corresponding to a latent word can be considered an observed sequence generated by an HDP-HSMM. 
If only a single sub-sequence of observations corresponds to a latent word,  a latent letter sequence could be sampled using an ordinal sampling procedure in the HDP-HSMM. However, observations containing the same latent word have to share the same latent letter sequence $w$. Therefore, latent letter sequences for observations with the same latent word are simultaneously sampled, given that they have the same latent letter sequence. 
We employ an approximate sampling procedure based on sampling importance resampling (SIR)~\cite{p_robotics}.

If we define the observations sharing the same latent word as $ \bm{y}^{1:k} = \{  \bm{y}^1, \bm{y}^2, \ldots , \bm{y}^k \}$ and the shared latent letter sequence as $w$, the posterior probability $P(w \mid \bm{y}^{1:k})$ becomes
\begin{align}
  P(w \mid \bm{y}^{1: k}) &\propto P(w) P(\bm{y}^{1:k} \mid
  w )\label{eq:6}\\
  & = \underbrace{P(w \mid \bm{y}^j)}_{sampling}\underbrace{ P(\bm{y}^j) \prod_{i \neq j}^{k} P( \bm{y}^i \mid w)}_{weight}\label{eq:8}
\end{align}
where $P(\bm{y}^j)$ in \aref{eq:8}, representing the likelihood of the observation, can be calculated using the backward filtering procedure in the HDP-HSMM. Probability $P(\bm{y}^i \mid w)$ can also be calculated in the same way as \aref{eq:5} if $w$ is given. The HDP-HSMM also provides a sampling procedure for $P(w \mid \bm{y}^j)$. Therefore, if we consider $P(w \mid \bm{y}^j)$ as the proposed distribution and $P(\bm{y}^j) \prod_{i \neq j}^{k} P( \bm{y}^i \mid w)$ as a weight, the SIR procedure can be employed~\cite{p_robotics}. 
Specifically, after a set of $w$ are sampled from the proposed distribution $P(w \mid \bm{y}^j)\ j=1, 2, \ldots ,k$, a final sample is drawn from the set with a probability proportional to each sample's weight. 
Using this procedure, the proposed model can approximately sample a latent letter sequence $w_i$ for the $i$-th latent word. 

\subsection{Sampling model parameters}
After sampling latent words $\{z_s\}$ for each observation data and sampling letter sequences for the latent words, other parameters can be updated.
Parameters of the language model, i.e., $\{\pi^{LM}_i\}$ and $\beta^{LM}$, can be updated on the basis of latent word sequences. 
Parameters of the word model, i.e., $\{\pi^{WM}_j\}$ and $\beta^{WM}$, can be updated on the basis of sampled letter sequences for latent words.
Parameters for the acoustic model, i.e., $\{\omega_j\}$ and $\{\theta_j\}$, can be updated if each hidden state $x_t$ is determined for each $y_t$. During the SIR process for sampling a letter sequence, $\{ \bar{w}^m_s  \}$ in Algorithm 1 are subsidiarily obtained. To accelerate the mixing rate, the subsidiary sampling results $\{ \bar{w}^m_s  \}$ obtained in the SIR are used for updating the acoustic model parameters.
These parameters can be sampled in the same way as the HDP-HSMM. For more details, we refer to the original paper in which the HDP-HSMM were introduced~\cite{johnson2013}.
Finally, the overall sampling procedure is obtained, as described in Algorithm~\ref{alg:bgibbs}.

\begin{algorithm}[tb]
\caption{Blocked Gibbs sampler for HDP-HLM}
\label{alg:bgibbs}
\begin{algorithmic}
\STATE Initialize all parameters.
\STATE Observe $M$ time series data $\{y^m_{1:T_m}\}_{m\in\{1, 2, \ldots, M\}}$. 
\REPEAT
\FOR{$m=1$ to $M$}
\STATE // Backward filtering procedure
\STATE For each ${i\in\{1, 2, \ldots, N\}}$, initialize messages $B_T(i)=1$.
\FOR{$t=T$ to $1$} 
\STATE For each ${i\in\{1, 2, \ldots, N\}}$, compute backward messages $B_{t-1}(i)$ and $B_{t-1}^*(i)$ using \aref{eq:1}--\aref{eq:4}.
\ENDFOR
\STATE // Forward sampling procedure
\STATE Initialize $s=1$ and $D^{\text{sum}}_s=0$
\WHILE{$D^{\text{sum}_s} < T_m$} 
\STATE // Sampling a super state representing a latent word
\STATE $z_s \sim p(z_s \mid y^m_{1:T_m}, z_{s-1}, F_{D^{\text{sum}}_s}=1 )$ 
\STATE // Sampling duration of the super state  
\STATE $D_s \sim p(D_s | z_s , F_{D^{\text{sum}}_s}=1  )$ 
\STATE  $D^{\text{sum}}_{s+1} \leftarrow D^{\text{sum}}_s + D_s$
\STATE $s \leftarrow s+1$
\ENDWHILE
\STATE $S^m \leftarrow s-1$
\STATE // Sampling a tentative latent letter sequences 
\FOR{$s=1$ to $S^m$} 
\STATE $ \bar{w}^m_s \sim P(w | y^m_{D^{\text{sum}}_{s-1}+1:D^{\text{sum}}_s}, \{\pi_j^{WM}, \omega_j, \theta_j \}_{j=1,2,\ldots,J})$ 
\ENDFOR
\ENDFOR
\STATE // Update model parameters
\STATE Sample acoustic model parameters $\{ \omega_j, \theta_j\}$ on the basis of tentatively sampled latent letter sequences $\{ \bar{w}^m_s\}$.
\STATE Sample language model parameter $\{\pi_i^{LM}\}, \beta^{LM}$ on the basis of sampled super states , i.e., latent words. 
\STATE Sample a word inventory $\{w_i\}_{i = 1, 2, \ldots, N}$ using SIR procedure (see \aref{eq:8}).
\STATE Sample a word model $\{\pi_i^{WM}\}, \beta^{WM}$ on the basis of sampled word inventory  $\{w_i\}_{i = 1, 2, \ldots, N}$. 
\UNTIL{a predetermined exit condition is satisfied.}
\end{algorithmic}
\end{algorithm}

\subsection{NPB-DAA}
Based on the generative model, HDP-HLM, and its inference algorithm shown in Algorithm~\ref{alg:bgibbs}, the proposed NPB-DAA is obtained, finally. 
By assuming HDP-HLM as a generative model of observed time series data, and by inferring latent variables of the model, we can analyze latent double articulation structure, i.e., hierarchically organized latent words and phonemes, of the data in an unsupervised manner. We call the novel unsupervised double articulation analyzer NPB-DAA.

\section{Experiment 1: Synthetic data}\label{sec5}
We conducted an experiment using a synthetic dataset that explicitly has a double articulation structure to validate our proposed method.

\subsection{Conditions}
To validate the ability of our proposed method to infer a latent double articulation structure in time series data, we applied the proposed NPB-DAA based on the HDP-HLM to synthetic time series data. 
The conventional DAA was employed as a comparative method.
The time series data are generated using five letters $ \{j\}_{j\in J} = \{1, 2, 3, 4, 5\}$ and four words $\{ w \}_{w\in W} =  \{[1, 3, 5], [3, 2], [4, 1, 5, 2], [1, 5]\}$ where $J$ is a set of letters and $W$ is a set of words. The  four words were generated randomly. 
The sequence $w_i = [w_{i1}, w_{i2}, \ldots , w_{iL_i}]$ represents a word that is generated by combining $\{w_{i1}, w_{i2}, \ldots , w_{iL_i} \}$ sequentially where $w_{ik}$ denotes the $k$-th letter of $w_i$. The durations of the letters were assumed to follow Poisson distributions and their parameters were drawn from a Gamma distribution whose parameters were $\alpha = 50, \beta = 10$. The emission distribution was assumed to be a Gaussian distribution whose parameters were $\mu = 5i, \sigma^2 \in \{ 0.1, 0.5, 1.0 \} $, where $i$ represents the index of latent letters. The variance of the emission distribution was changed in stages, and the inference results were compared.  Forty time series data items were generated from 20 types of latent word sequences. Sixteen of them were pairs of words in $W$, e.g., $([1, 3, 5], [1, 5] )$ , and $([3 ,2], [3 ,2] )$. Four of them were three-word sentences, e.g., $([3, 2], [1, 3, 5], [1, 5]  )$.  A sequence of latent words is represented by $(w_1, w_2, \ldots , w_n)$.  Two observations were generated from each word sequence.

We set the parameters of the NPB-DAA as follows: the hyperparameters for the latent language model were $\gamma^{LM} = 10.0, \alpha^{LM} = 10.0$, and the maximum number of words was six for weak-limit approximation. The hyperparameters for the latent word model were $\gamma^{WM} = 10.0, \alpha^{WM} = 10.0$, and the maximum number of letters was seven for weak-limit approximation. The hyperparameters of the duration distributions were set to $\alpha = 50$ and $\beta = 10$, and those of the emission distributions were set to $\mu_0 = 0,
\sigma^2_0 = 1.0, \kappa_0 = 0.01, \nu_0 = 1$. The Gibbs Sampling procedure was iterated 100 times.

For the conventional DAA, we set the hyperparameters of the sticky HDP-HMM to be as similar to those of the NPB-DAA as possible. In this condition, the latticelm software\footnote{latticelm: http://www.phontron.com/latticelm/index.html} developed by Neubig et al. was used for NPYLM. The hyperparameters of the NPYLM used in the conventional DAA were set to $\alpha = 0.1$ and $d = 0.1$.

The hyperparameters in the NPB-DAA were heuristically given in a top-down manner by referring to the size of the state space and the approximate duration of a phoneme. Those of the Pitman-Yor language model were set to the default values of the software.

\subsection{Results}

\begin{figure}[tb]
    \centering
    \includegraphics[width=0.95\linewidth]{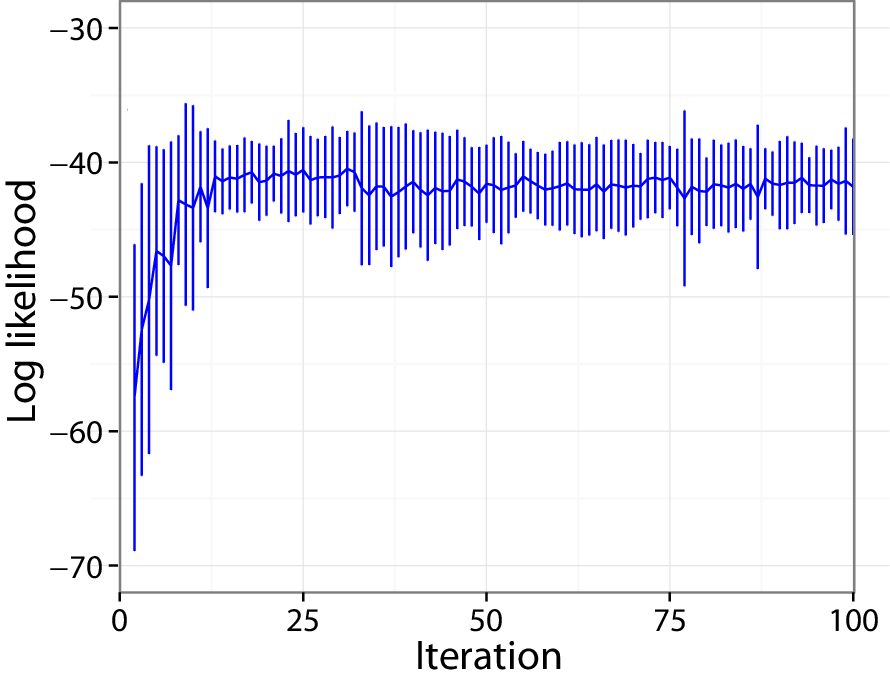}
  \caption{Log-likelihood profile through Gibbs sampling ($\sigma^2=1.0$)}
  \label{fig:llk}
\end{figure}

The average log-likelihood is shown in Fig.~\ref{fig:llk}, where error bars represent the standard deviation of 30 trials. These results show that the proposed inference procedure worked appropriately, gradually sampling more probable latent variables as the iterations increased.

In contrast with ordinal speech recognition tasks, the target task (language acquisition and double articulation analysis) is an unsupervised learning task. Specifically, it is a clustering task. Therefore, it is difficult to evaluate the methods' performance from the viewpoint of precision and recall because the estimated index of a cluster and the label corresponding to the ground truth data are usually different. We evaluated the obtained result using the adjusted rand index (ARI), which quantifies the performance of a clustering task~~\cite{ari}.
If all data items are clustered randomly or only to one cluster, the ARI becomes $0$. By contrast, if the results of clustering are the same as those of the ground truth data, the ARI becomes $1$.

Table~\ref{exp1_ari_letter} shows the ARI for the estimated latent letters. The ARI for estimated latent letters shows how accurately each method estimated latent letters, which correspond to phonemes in speech signals.
Table~\ref{exp1_ari_word} shows the ARI for estimated latent words. The ARI for estimated latent words shows how accurately each method estimated latent letters, which correspond to words in speech signals.
In both tables, each column shows ARIs for different {$\sigma^2$}.
A higher ARI implies more accurate estimation of the latent variables.

\begin{figure}[tb]
    \centering
    \includegraphics[width=0.9\linewidth]{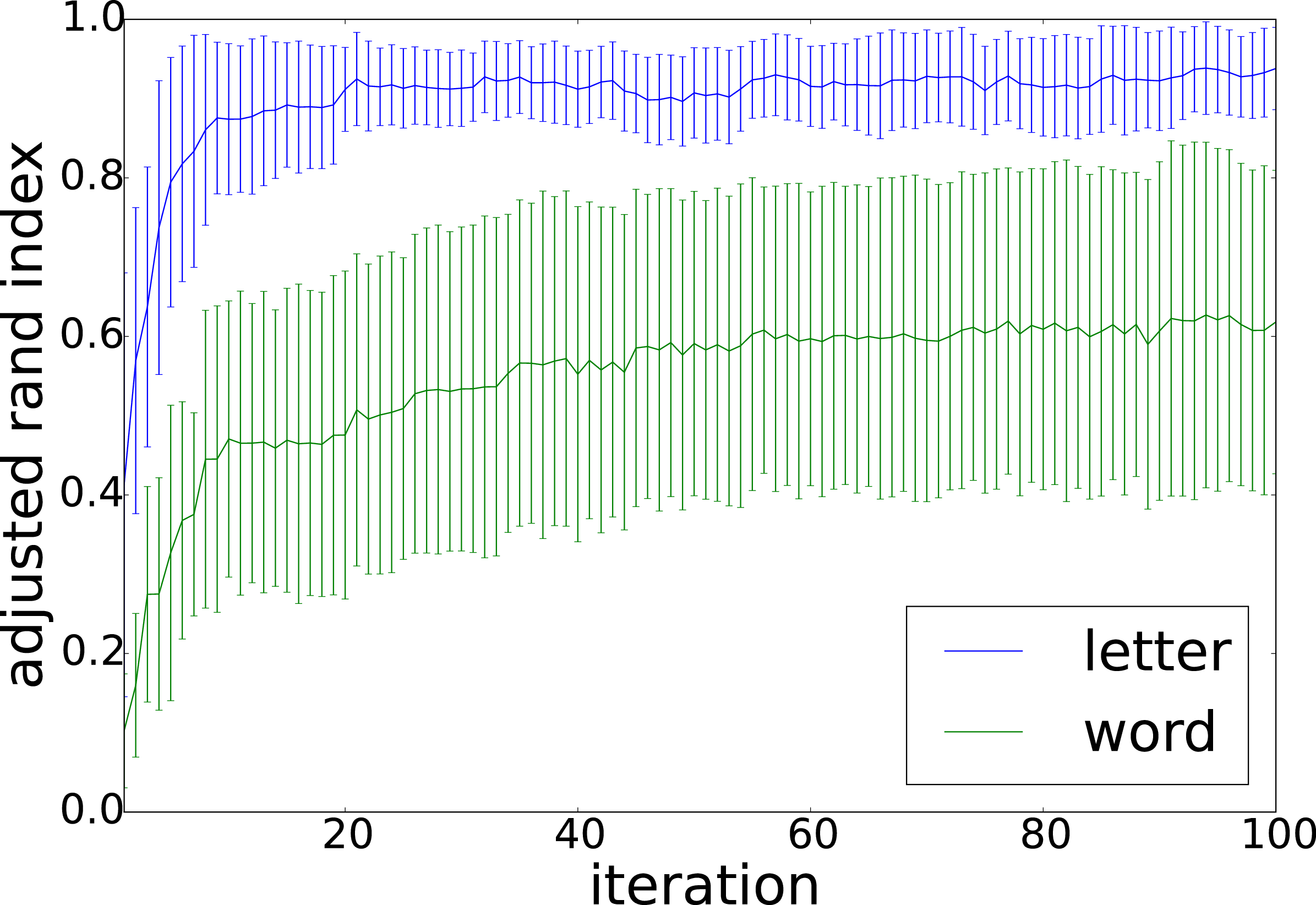}
  \caption{ARI profile through Gibbs sampling ($\sigma^2=1.0$)}
  \label{fig:aritime}
\end{figure}

Although the ARI for the latent letters obtained by conventional DAA decreases when the variance $\sigma^2$ increases, that of  NPB-DAA did not decrease as much. As the ARIs for latent words show, the performance of word segmentation by conventional DAA was poor, even when the ARI for latent letters was larger than $0.8$. In contrast, the ARI for latent words estimated by NPB-DAA was over $0.5$ in all conditions.
This shows that the NPB-DAA can mitigate the ill effects of phoneme recognition errors in the word segmentation task, and obtained knowledge about words can improve phoneme recognition performance by using contextual information. 
Fig.~\ref{fig:aritime} shows the change in ARI through iterations in the case of $\sigma^2 = 1.0$. This shows that the ARI also increased gradually while log likelihood increases, as in Fig.~\ref{fig:llk}. 
These results suggest that the NPB-DAA is an appropriate generative model because better word segmentation performance corresponded to higher likelihood of the model.

To check the effects of the limit on weak-limit approximation, we ran an experiment where the maximum number of letters was 20 for weak-limit approximation. The ARI for the estimated latent words were $\{ 0.682, 0.650, 0.604 \}$, those for estimated latent letters were $\{ 0.967, 0.899, 0.878 \}$, and the estimated number of latent letters were $\{ 5.6, 6.3, 6.6 \}$ on average for $\sigma^2 =\{0.1, 0.5, 1.0\}$. This result shows that our model can work appropriately to estimate the number of latent states owing to the nature of Bayesian nonparametrics when the limit is sufficiently large.

\begin{figure}[tb!p]
    \centering
    \includegraphics[width=\linewidth] {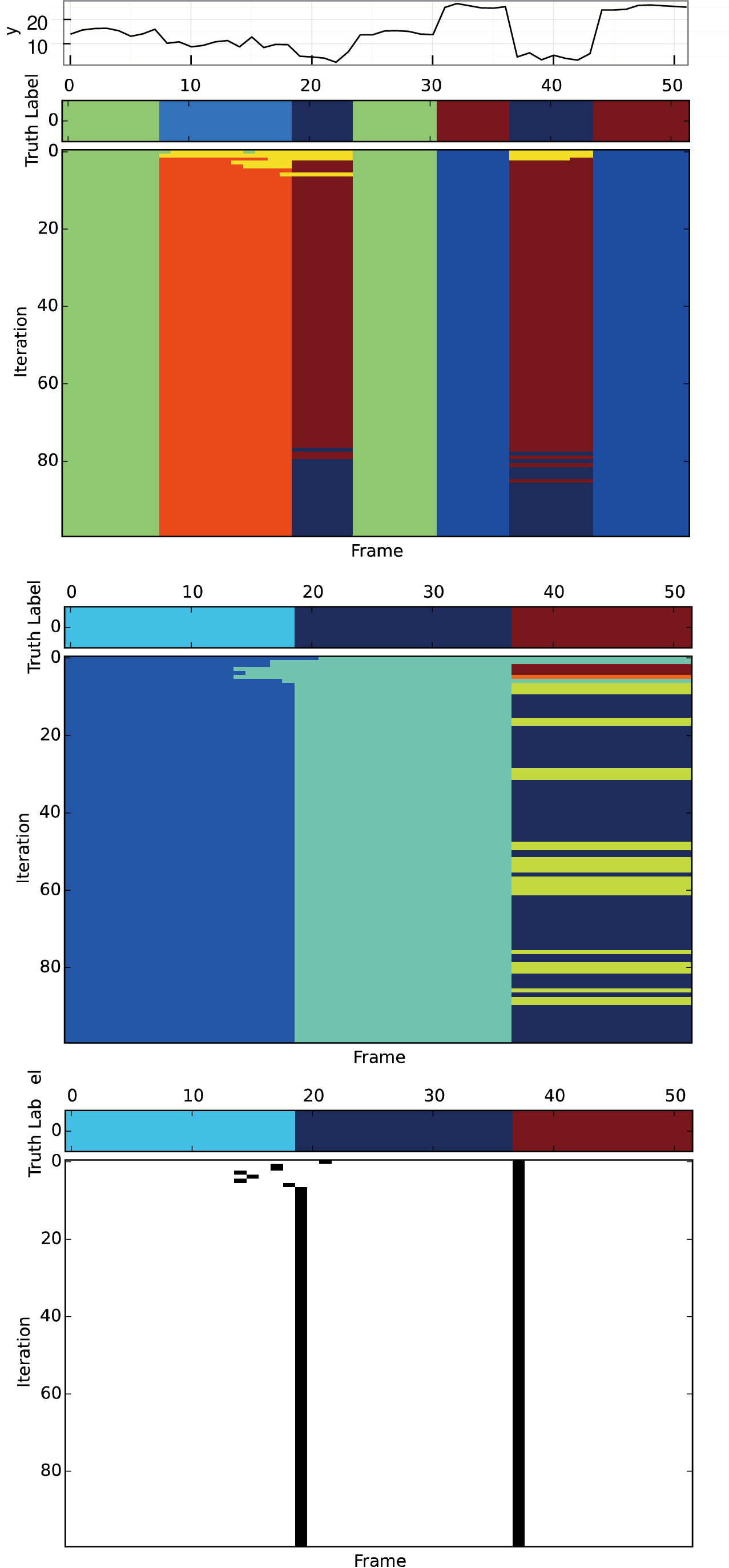}
%
  \caption{Example of inference results for sample data $([3, 2], [1, 3, 5], [1, 5])$ and $ \sigma ^2 = 1.0 $: (top) observation data, (upper middle) latent letters, (lower middle) latent words, and (bottom) the boundaries of latent words. Different colors denote different states.}
  \label{fig:state1}
\end{figure}

An example of estimated latent variables is shown in Fig.~\ref{fig:state1}, which shows the results for time series data generated from the latent word sequence $([3, 2], [1, 3, 5], [1, 5])$. The input time series data is shown at very top of the figure. The top of each panel shows the true latent letters or latent words, whereas the panel beneath shows the inferred results. The vertical axes represent the iteration of the Gibbs sampling. In Fig.~\ref{fig:state1}, the figure in the middle shows a latent word sequence estimated using the proposed method, and the figure at the bottom shows the estimated boundaries of the latent words. These results show that the inference procedure works consistently and can estimate an adequate boundary for the latent words given the data.

\begin{table}[bt]
\caption{ARI for estimated latent letters}
\label{exp1_ari_letter}
\begin{center}
\begin{tabular}{|c|c|c|c|}
\hline
$\sigma^2$ & 0.1 & 0.5 & 1.0 \\
\hline
\shortstack{Conventional DAA\\ (sticky HDP-HMM)} & 0.845 & 0.832 & 0.649 \\
\hline
NPB-DAA & {\bf 0.984} & {\bf 0.895} & {\bf 0.938} \\
\hline
\end{tabular}
\end{center}
%
\caption{ARI for estimated latent words}
\label{exp1_ari_word}
\begin{center}
\begin{tabular}{|c|c|c|c|}
\hline
$\sigma^2$ & $0.1$ & $0.5$ & $ 1.0 $ \\
\hline
\shortstack{Conventional DAA\\ (sticky HDP-HMM + NPYLM) } & 0.122 & 0.107 & 0.125 \\
\hline
NPB-DAA & {\bf 0.594} & {\bf 0.509} & {\bf 0.618} \\
\hline
\end{tabular}
\end{center}
\end{table}

These results show that the proposed method is a more effective machine learning method for estimating a latent double articulation structure embedded in time series data.

\section{Experiment 2: Continuous Japanese Vowel Speech Signal}\label{sec6}
In the second experiment, we evaluated our proposed method using Japanese vowel speech signals to test the applicability of the proposed method to actual human continuous speech signal. 

\subsection{Conditions}
We prepared four datasets. Each dataset corresponds to a speaker, and consisted of 60 audio data items. We asked two male and two female Japanese speakers to read 30 artificial sentences aloud two times at a natural speed, and recorded his/her voice. 
The 30 sentences were prepared using five words \{aioi, aue, ao, ie, uo\}, which consisted of five Japanese vowels \{a, i, u, e, o\} representing {\{}{\textipa{\"a, i, W\super B, \|`e, \|`o}}{\}} in phonetic symbols respectively. By reordering the five words, we prepared 25 two-word sentences, e.g., ``ao aioi,'' ``uo aue,'' and ``aioi aioi,'' and five three-word sentences, i.e., ``uo aue ie,'' ``ie ie uo,'' ``aue ao ie,'' ``ao ie ao,'' and ``aioi uo ie.'' The set of two-word sentences consisted of all types of word pairs ($5 \times 5 =25$). The set of three-word sentences were generated randomly.


The recorded data were encoded into $13$-dimensional mel-frequency cepstrum coefficient (MFCC) time series data using the HMM Toolkit (HTK)\footnote{Hidden Markov Model Toolkit: http://htk.eng.cam.ac.uk/}. The frame size and shift were set to $25$ and $10$ ms, respectively. Twelve-dimensional MFCC data was obtained as input data by eliminating power information from the original 13-dimensional MFCC data. As a result, 12-dimensional time series data at a frame rate of $100$ Hz were obtained.


The hyperparameters for the latent language model were set to $\gamma^{LM} = 10.0$ and $\alpha^{LM} = 10.0$, and the maximum number of words was set to seven for weak-limit approximation. The hyperparameters for the latent word model were $\gamma^{WM} = 10.0$ and $\alpha^{WM} = 10.0$, and the maximum number of letters was seven for weak-limit approximation. 
The hyperparameters of the duration distributions were set to $\alpha = 200$ and $\beta = 10$, and those of the emission distributions were set to $\mu_0 = 0, \sigma^2_0 = 1.0, \kappa_0 = 0.01,$ and  $\nu_0 =  17 = ($dimension$ + 5)$.

For the conventional DAA, we set the hyperparameters of the sticky HDP-HMM to be as similar to those of the NPB-DAA as possible. The hyperparameters for the NPYLM used in the conventional DAA were set to $\alpha = 0.1$ and $d = 0.1$. The Gibbs sampling procedure was iterated 100 times. With different random number seeds, 20 trials were performed. 

The parameters in the NPB-DAA were given in a top-down manner heuristically by referring to the size of the state space and the approximate duration of a phoneme. Those of the Pitman-Yor language model were set to the default values of the software.

As a baseline method, we employed an open-source continuous speech recognition engine, Julius,\footnote{Open-Source Large Vocabulary CSR Engine Julius: http://julius.sourceforge.jp/. The Linux binary \textsf{dictation-kit-v4.3.1-linux.tgz} was used in this experiment. The software encodes the recorded data into 36-dimensional MFCC data including dynamic features and uses them for speech recognition. } which is widely used in Japanese speech recognition tasks. Julius's acoustic model is trained by using a large number of speech data in a supervised manner. We prepared four conditions for Julius. The first one was called ``Julius (phoneme + NPYLM).'' In this condition, we used  Julius as a phoneme recognition system by preparing a phoneme dictionary containing five Japanese vowels \{a, i, u, e, o\}. Moreover, Julius's dictionary also contains silB and silE to represent silence due to system requirements. After encoding continuous speech signals into phoneme sequences using Julius as a phoneme recognizer, unsupervised morphological analysis based on the NPYLM was conducted to discover words and a language model.  The second condition was called  ``Julius (phoneme + latticelm).'' In this condition, we also used latticelm, which is an unsupervised morphological analyzer for lattice output from an ASR system. The method was proposed by Neubig et al. as an extension of Mochihashi's NPYLM~\cite{Neubig2012}. In this condition, the latticelm software was used too.    

In the third and fourth conditions, called ``Julius (monophone + word dictionary)'' and ``Julius (triphone + word dictionary),'' respectively, we prepared a complete word dictionary that contained all of the words that appeared in the target speech signal, i.e.,\{aioi, aue, ao, ie, uo\}, for Julius. This condition provides almost an upper bound for the performance of our task.
Except for in ``Julius (triphone + word dictionary),'' Julius uses a monophone-based acoustic model contained in the dictation kit. The acoustic model is trained in a supervised manner using a large number of labeled speech data.  ``Julius (triphone + word dictionary)'' used a triphone-based acoustic model for comparison. 

\subsection{Results}

 We provided word and letter ground truth labels to all frames of the speech signal data and evaluated the relationship between the truth labels and estimated latent letter and word indices.

\begin{table}[bt]
\caption{ARI for estimated latent letters and words}
\label{tbl:exp2_ari}
\begin{center}
\begin{tabular}{|c||c|c||c|c|}
\hline
Method & \shortstack{Letter ARI} & \shortstack{Word ARI} &AM & LM \\
\hline\hline
NPB-DAA  (MAP)& {\bf 0.596} & {\bf 0.529}  &&  \\
\hline
NPB-DAA  & {0.561} & {0.401} &&  \\
\hline
Conventional DAA  & {0.590} & {0.090}  && \\
\hline\hline
\shortstack{Julius (phoneme dictionary\\ + NPYLM)} & {0.486} & {0.297} & \checkmark &\\
\hline
\shortstack{Julius (phoneme dictionary\\ + latticelm)} & {0.554} & {0.337} & \checkmark &\\
\hline
\hline\hline
\shortstack{Julius (monophone\\ + word dictionary)} & {0.586} &  {0.487} & \checkmark & \checkmark\\
\hline
\shortstack{Julius (triphone\\ + word dictionary)} & {0.548} &  {0.616} & \checkmark &  \checkmark \\
\hline
\end{tabular}
\end{center}
\end{table}

The results are shown in Table~\ref{tbl:exp2_ari}.  Check marks in the AM and LM columns indicate that the method used a pretrained acoustic model (AM) and the given true language model (LM), respectively. 
Letter ARI shows the ARI of phoneme clustering. A high Letter ARI means more accurate phoneme acquisition and recognition. Word ARI shows the ARI of word clustering. A higher Word ARI means more accurate word discovery and recognition. Each row corresponds to each method explained in the conditions.
The results of ``NPB-DAA'' and ``Conventional DAA'' show the  ARI averaged over 20 trials. In contrast, ``NPB-DAA (MAP)'' obtained the maximum a posteriori probability (MAP) of the 20 trials.
An advantage of the NPB-DAA is that the method can calculate the posterior probability of a given dataset after the learning phase because the NPB-DAA is derived from a generative model, i.e., HDP-HLM, which integrates the language and acoustic models. In contrast with the conventional DAA and similar methods that do not have appropriate generative models, the NPB-DAA can obtain an appropriate learning result by referring to the probability. The rows with MAP in Table~\ref{tbl:exp2_ari} show that this probability is an adequate criterion for selecting a learning result.

The results show that the ``NPB-DAA (MAP)'' outperformed not only the conventional DAA but also Julius-based word discovery systems whose acoustic models were trained in supervised manner.
One reason is that the acoustic models of the DAAs were trained only from one participant's speech signals, in contrast, Julius's acoustic model was trained by the speech signals of many speakers. 
In other words, NPB-DAA acquired speaker-dependent acoustic model in contrast with that Julius used speaker-independent acoustic model.
This adaptation of acoustic model to the speaker must have increased the NPB-DAA's performance.  

The results show that a naive application of the NPYLM to recognized phoneme sequences results in poor word acquisition performance, especially in conventional DAA. Because the theory of the NPYLM does not presume that letter sequences have recognition errors, the existence of phoneme recognition error deteriorates word segmentation performance. The methods that simply apply an NPYLM  to obtained phoneme sequences, i.e., the conventional DAA and Julius (phoneme dictionary + NPYLM), output bad results in the word ARI  compared with those of the letter ARI. However, latticelm, which presumes phoneme recognition errors to some extent, could not dramatically improve the performance of word acquisition in our experimental setting. 

In contrast, ``Julius (triphone + word dictionary)'' improved its word ARI performance with respect to letter ARI performance. ``Julius (monophone + word dictionary)'' also kept its performance high with respect to the word recognition task compared with the phoneme recognition task.
We note that the word error rate was 32.8\% and the phoneme error rate was 28.1\% in Julius (monophone + word dictionary).

In the research field of ASR, it is widely known that a good language model improves word and phoneme recognition performance. The NPB-DAA could not improve the performance of word ARI with respect to letter ARI performance. However, it obtained an adequate language model and prevented the score of the word ARI from becoming far worse than that of the letter ARI. To achieve such an error-proof word acquisition, the direct inference of latent words are important in NPB-DAA. In the inference procedure described in Section~\ref{sec3}, latent words are sampled directly without sampling latent letters while marginalizing all possible latent letter sequences.  This achieves an effect similar to that of a given language model in the inference process

Typical examples of the estimation results are shown in Table~\ref{tbl:DAAresult} for NPB-DAA and conventional DAA. Each number in parentheses  represents an estimated phoneme label, each space represents a phoneme boundary, each number in bold style represents a sampled index of a word, and ``$/$'' represents a boundary between successive words. For example, ``ao ie'' was divided into two words, i.e., ``5 0 1'' and ``6 3 4 6,'' in the NPB-DAA results, and their word indices were 3 and 4. In Table~\ref{tbl:DAAresult}, the sampled letters corresponding to the word ``ie'' are underlined. Although conventional DAA could not estimate ``ie'' as a single word, the NPB-DAA could estimate ``ie'' to be a single word: ``4.'' In the conventional DAA results, several phoneme recognition errors can be found. The errors completely deteriorated the following chunking process, i.e., unsupervised morphological analysis using a NPYLM, as past research has frequently pointed out.
As shown in Table~\ref{tbl:DAAresult},  NPB-DAA had some phoneme recognition errors. However, in the NPB-DAA, latent words are sampled on the basis of the marginalized phoneme distribution before sampling concrete phoneme sequences. This property of the sampling procedure seemed to improve the performance of NPB-DAA.

\begin{table*}[tb]
\caption{Example word discovery results }
\label{tbl:DAAresult}
\begin{center}
\begin{tabular}{|c|c|c|}
\hline
Vowel sequence & Estimated NPB-DAA results  &  Estimated conventional DAA results\\
\hline \hline
ao ie &	{\bf 3} (5 0 1) / \uline{{\bf 4} (6 3 4 6)} 											&	{\bf 226} (2 0 3 4  \uline{1 5 4 1})	\\	
ao ie ao	& {\bf 3} (5 0 1) /  \uline{{\bf 4} (6 3 4 6)} / {\bf 3} (5 0 1) / {\bf 0} (6 4 6) 	&	{\bf 494} (3) / {\bf 675} (2 3 0) / {\bf 374} ( \uline{1 5 4 1} 2 0 1)	\\	
aue ie	& {\bf 6} (6 5 1 2 6 4) /  \uline{{\bf 4} (6 3 4 6)} 										&	{\bf 329} (2 3 8 4  \uline{5 4 1})	\\	
ie ie	&  \uline{{\bf 4} (6 3 4 6)} /  \uline{{\bf 4} (6 3 4 6)} 								&	{\bf 389} ( \uline{5 4 1 4}  \uline{1 5 4 1})	\\	
ie uo	&  \uline{{\bf 4} (6 3 4 6)} / {\bf 5} (5 1 2) / {\bf 3} (5 0 1)  				&	{\bf 401} ( \uline{5 4 1} 8 0 1)	\\	
ie aioi &	 \uline{{\bf 4} (6 3 4 6)} / {\bf 1} (5 6 4 6 3 6 1) /  {\bf 4} (6 3 4 6)			 	&	{\bf 813} ( \uline{5 4 1} 2 4 5) / {\bf 832} (4 3 0 3 4 5 1)	\\	
\hline
\end{tabular}
\end{center}
\end{table*}

\begin{figure}[tb]
    \centering
    \includegraphics[width=\linewidth]{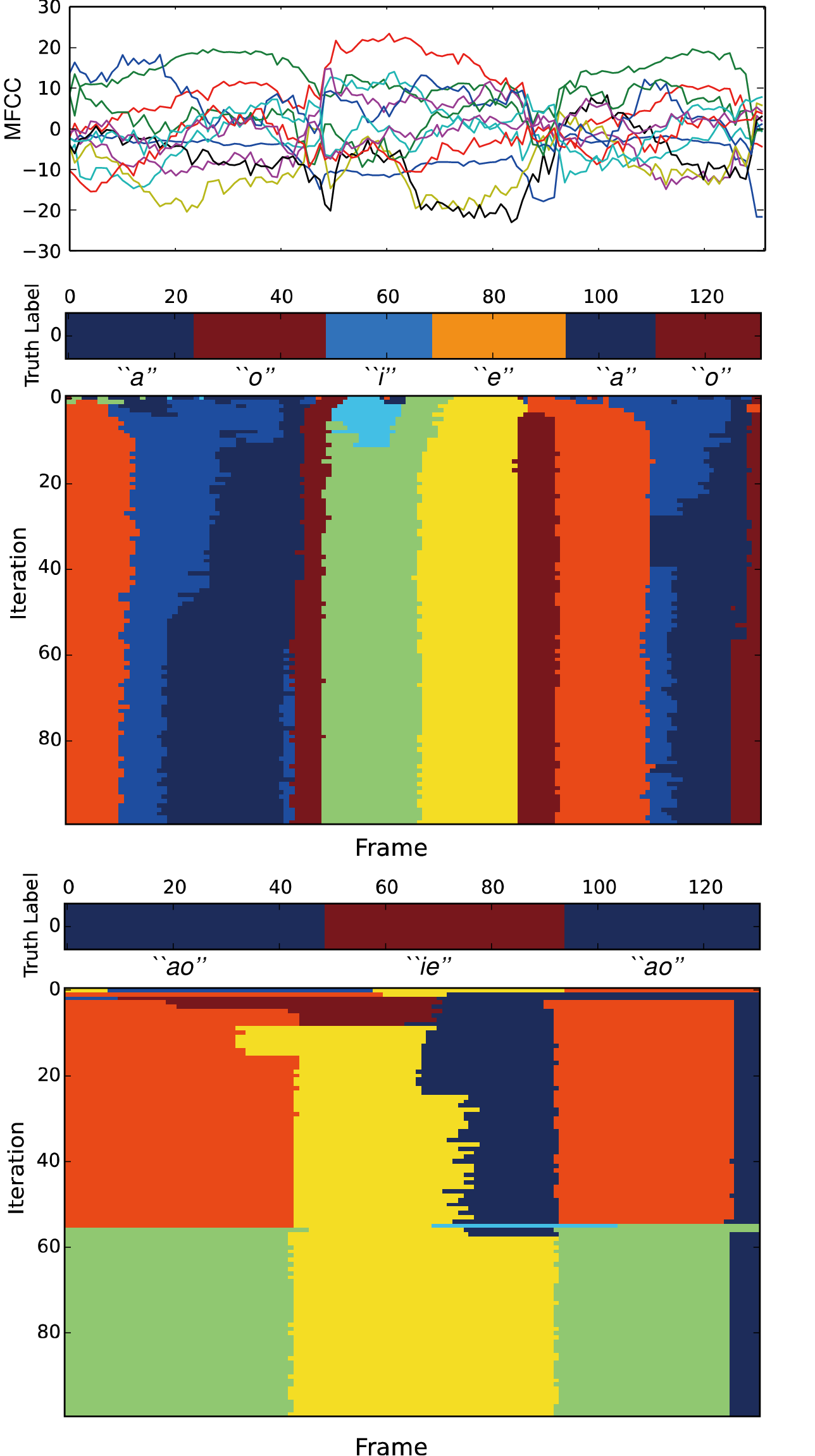}
  \caption{Example of inference results for ``ao ie ao.'' MFCC feature vectors are plotted in the top panel. The middle and bottom panels show the inference results of latent letters and latent words, respectively. Different colors denotes different states. }
 \label{fig:npb_example_inference}
\end{figure}

An example of the estimated latent variables is shown in Fig.~\ref{fig:npb_example_inference}, which shows the results for time series data corresponding to a vowel sequence, ``ao ie ao.'' The input time series data, i.e., 12-dimensional MFCC time series data, are shown at the top of the figures. The middle and the bottom figures show the inference process. The top of each figure shows the true latent letters or latent words, whereas the bottom shows the inferred result. The vertical axes represent the number of Gibbs sampling iterations. This shows that the inference procedure worked for human vowel sequence data, and could estimate an adequate unit for each word. 

Let us further examine the characteristics of the segmentation results of the NPB-DAA. Table~\ref{tbl:DAAresult} shows that some of the estimated latent words have a latent letter ``6'' at their head or tail. The latent letter ``6'' represents silence observed during the transition from one vowel to another. Silence in speech signals and the transitional sounds observed between two phonemes were treated in the same manner as other uttered sounds in our model. The question of whether such signals should be treated in the same way as other sounds in a generative model calls for further investigation. In our model, a phoneme is simply represented by a single Gaussian distribution, although many past speech recognition systems assign a richer structure to a phoneme, e.g., a three-state left-to-right HMM with GMM emission distributions. There is room for investigating whether a phoneme model, i.e., a latent letter, should itself have a more complex structure, or if a double articulation hierarchy is sufficient from the viewpoint of unsupervised word discovery tasks.

An interesting result that represents a characteristic of the NPB-DAA is the latent word ``{\bf 4} (6 3 4 6)'' estimated at the end of ``ie aioi.'' The speech signals corresponding to this ``{\bf 4}'' were a kind of transitional sound observed following ``aioi.'' The NPB-DAA directly inferred the latent word by marginalizing latent letters. In this case, it seems that ``{\bf 4}'' was more likely than other latent words, and the NPB-DAA hence generated this result. This can be regarded as a side effect of our approach, i.e., the marginalization of latent letter sequences in a latent word. We are confident that the marginalization of latent letters and the direct inference of word sequences are important to improving the performance of the unsupervised word segmentation of continuous speech signals, but there is room to consider this side effect.


Note that the NPB-DAA performed unsupervised word discovery under the condition that the training data consisted of speech signals uttered by one speaker, in contrast with Julius, whose acoustic model was trained using many speakers' speech signals. Speaker-independent, unsupervised word discovery from continuous speech signals remains a challenging problem because the acoustic features of phonemes heavily depend on the speaker. When we gave four speakers' speech signals to the NPB-DAA at the same time, the Letter ARI and the Word ARI decreased to $0.297$ and $0.104$, respectively. By contrast, those produced by Julius with a triphone acoustic model and a true word dictionary were $0.552$ and $0.599$, respectively. 
In the experiment, 120 audio data items that were recorded by asking two male and two female Japanese speakers to read 30 artificial sentences were used, i.e., a half of the data items used in the main experiment due to computational cost. It was observed that speaker ``dependent'' phoneme models were obtained by the NPB-DAA, i.e., speech signals representing the same phoneme uttered by deferent persons tended to be clustered to different latent letters. To develop a machine learning method that enables a robot to obtain language and acoustic models independent of speakers, or automatically adapting to different speakers is one of our future challenges.

\section{Conclusion}\label{sec7}
In this paper, we proposed NPB-DAA for direct and simultaneous acquisition of language and acoustic models from continuous speech signals in an unsupervised manner. For this purpose, we proposed an integrative generative model called the HDP-HLM by extending HDP-HSMM. Based on the generative model, we derived an inference procedure by extending the blocked Gibbs sampler originally proposed for HDP-HSMM. The method is expected to enable a developmental robot to simultaneously obtain language and acoustic models directly from continuous speech signals. To evaluate the performance of the proposed method, two experiments were performed. In the first experiment, the proposed method was applied to synthetic data, and it was shown that the method can successfully infer latent words embedded in time series data in an unsupervised manner. In the second experiment, we applied the proposed method to actual human Japanese vowel sequences. The result showed that the proposed method outperformed a conventional two-stage sequential method, conventional DAA, and a baseline ASR method.

One of the most important challenges in our future work is to achieve complete human language acquisition from speech signals. We did not achieve complete language acquisition from speech signals that includes consonants as well as vowels in this study. Language acquisition from more natural speech signals like child-directed speech by human parents are also part of our future work. To achieve these aims, we still have two main problems: feature extraction and computational cost.

 To address these problems, more sophisticated feature extraction methods are needed. Deep learning has gained attention recently because of its impressive feature extraction performance. Integrating a deep learning method into the NPB-DAA should improve its performance.

Computational cost is another problem. Even though the size of the
dataset used in the Experiment 2 was very small, it took approximately 240
minutes for 100 iterations using an Intel Xeon CPU E5-2650 v2 2.60 GHz, 8 cores $\times$ 16
CPU. In particular, the computational cost of the blocked Gibbs sampler was $O(TL_{max}d_{max}^3 N_{max}^2)$, where $L_{max}$ is the maximum number of latent letters for a word, $d_{max}$ is the maximum duration of a word, and $N_{max}$is the maximum number of words. 
To apply the proposed method to a larger dataset, improving its computational cost will be necessary. 

Currently, the accuracy of the language acquisition is still limited, as shown in Table~\ref{tbl:exp2_ari}. In this paper, we focused on a language acquisition method based on distributional cues and proposed a mathematical model for language acquisition. Obviously, distributional cues are not enough for more accurate language acquisition. As suggested by several computational and robotic studies, making use of co-occurrence cues improves the accuracy of language acquisition~\cite{Nakamura2014, Akira2015, Taguchi2011}. The proposed HDP-HLM is a fully probabilistic generative model. Therefore, introducing other factors into consideration is relatively easier than for other heuristic models. This is also advantage of our approach. Combining prosodic and co-occurrence cues into the NPB-DAA, and obtaining a more accurate and more plausible constructive developmental language acquisition model is also a direction for future research. 

\bibliographystyle{IEEEtran}
\bibliography{aistats14}

ad
\end{document}